# PGF-Net: A Progressive Gated-Fusion Framework for Efficient Multimodal Sentiment Analysis

Bin Wen
School of Computer Sciences
Universiti Sains Malaysia
Penang, Malaysia
wenbin@student.usm.my

Tien-Ping Tan
School of Computer Sciences
Universiti Sains Malaysia
Penang, Malaysia
tienping@usm.my

*Abstract*—We introduce PGF-Net (Progressive Gated-Fusion Network), a novel deep learning framework designed for efficient and interpretable multimodal sentiment analysis. Our framework incorporates three primary innovations. Firstly, we propose a Progressive Intra-Layer Fusion paradigm, where a Cross-Attention mechanism empowers the textual representation to dynamically query and integrate non-linguistic features from audio and visual streams within the deep layers of a Transformer encoder. This enables a deeper, context-dependent fusion process. Secondly, the model incorporates an Adaptive Gated Arbitration mechanism, which acts as a dynamic controller to balance the original linguistic information against the newly fused multimodal context, ensuring stable and meaningful integration while preventing noise from overwhelming the signal. Lastly, a hybrid Parameter-Efficient Fine-Tuning (PEFT) strategy is employed, synergistically combining global adaptation via LoRA with local refinement through Post-Fusion Adapters. This significantly reduces trainable parameters, making the model lightweight and suitable for resource-limited scenarios. These innovations are integrated into a hierarchical encoder architecture, enabling PGF-Net to perform deep, dynamic, and interpretable multimodal sentiment analysis while maintaining exceptional parameter efficiency. Experimental results on MOSI dataset demonstrate that our proposed PGF-Net achieves state-of-the-art performance, with a Mean Absolute Error (MAE) of 0.691 and an F1-Score of 86.9%. Notably, our model achieves these results with only 3.09M trainable parameters, showcasing a superior balance between performance and computational efficiency.

## I. Introduction

The exponential growth of social media and video-sharing platforms has ushered in an unprecedented multimodal era. Human communication is no longer confined to text but is conveyed through a complex interplay of language, vocal tones, facial expressions, and physical gestures (Baltrušaitis et al., 2018). This inherently multidimensional nature of communication presents a significant challenge for artificial intelligence: to achieve genuine human-computer empathy and understanding, machines must be capable of parsing these multiple information channels. Consequently, Multimodal Sentiment Analysis (MSA) has emerged as a critical and challenging frontier in affective computing and artificial intelligence (Liu, 2012). Its core objective is to design computational models that can effectively fuse heterogeneous data streams—such as text, audio, and vision—to achieve a more accurate and robust identification of sentiment states than any single modality can provide alone. For instance, the textual statement "That's just great" can only be correctly interpreted as negative sentiment when combined with a sarcastic tone of voice and a frustrated facial expression, highlighting the indispensable need for multimodal information fusion.

In recent years, the advent of Large Language Models (LLMs) such as BERT (Devlin et al., 2019) has revolutionized the landscape of natural language processing and has been rapidly adopted for multimodal tasks. Researchers have achieved significant progress in MSA by integrating visual and acoustic features with powerful textual representations, leading to a series of advanced models like MulT (Tsai et al., 2019) and MAG-BERT (Sun et al., 2020). However, the paradigm of full fine-tuning large models faces two major bottlenecks: firstly, the immense computational and storage overhead makes deployment in resource-constrained environments challenging; secondly, fine-tuning all parameters on limited downstream task data can lead to the model forgetting the general knowledge acquired during pre-training, a phenomenon known as "catastrophic forgetting". To address these challenges, Parameter-Efficient Fine-Tuning (PEFT) techniques have emerged. Mainstream methods such as Adapters (Houlsby et al., 2019) and (Hu et al., 2022) have successfully reduced training costs by freezing the majority of pre-trained parameters and only training a small number of additional modules, all while maintaining high performance.

Against this backdrop, combining PEFT principles with the Mixture-of-Experts (MoE) model (Shazeer et al., 2017) offers a promising avenue for constructing efficient multimodal systems. Our work builds upon the Mixture of Multimodal Adapters (MMA) (Chen et al., 2025), a pioneering framework that leverages lightweight Adapters as experts and employs a gating network to route information from different modalities. While this design achieves remarkable parameter efficiency, our in-depth analysis reveals three core limitations that can be significantly improved:

1. **Static Feature Fusion Mechanism:** The initial fusion relies on a fixed, parameter-free attention mechanism, which is incapable of dynamically adjusting the contribution of each modality based on the context. This may lead to the loss of critical dynamic correlations at the very first stage of information fusion.

2. **Flat Routing Structure:** Its single gating network is required to make a complex, one-time decision among all expert groups (e.g., text, audio, and vision). This "flat" routing approach lacks a clear logical hierarchy and renders the model's decision-making process an uninterpretable "black box".

3. **Homogeneous Expert Design:** All experts (Adapters) within the original MMA framework share an identical internal architecture. This overlooks the fundamental differences in the statistical properties and information representation across different modalities, thereby limiting the specialized processing capability for modality-specific features.

To systematically address these limitations, we propose the Progressive Gated-Fusion Network (PGF-Net), a novel framework that fundamentally shifts the paradigm from parallel routing to serial, intra-layer fusion. Instead of making a single, complex routing decision, PGF-Net introduces three synergistic innovations that are progressively applied within the deep layers of the encoder: 1) a Progressive Intra-Layer Fusion paradigm using Cross-Attention to dynamically extract non-linguistic features; 2) an Adaptive Gated Arbitration mechanism to intelligently balance and integrate multimodal information streams; and 3) a Post-Fusion Adapter to refine the newly fused representations, all within a parameter-efficient design.

The main contributions of this paper can be summarized as follows:

- We propose PGF-Net, a novel parameter-efficient framework centered on a Progressive Intra-Layer Fusion paradigm, which achieves deeper and more dynamic integration of multimodal signals compared to traditional parallel-expert models.

- We design and validate a cohesive fusion module that synergistically combines Cross-Attention for information extraction, an Adaptive Gate for information arbitration, and a Post-Fusion Adapter for refinement, demonstrably enhancing fusion quality.

- We showcase the effectiveness of a hybrid PEFT strategy, where global LoRA adaptation and local Adapter refinement work in tandem to achieve state-of-the-art performance with significantly fewer trainable parameters.

## II. RELATED WORKS

This chapter provides a comprehensive review of the key research areas that form the foundation of our work. We first discuss the evolution of Multimodal Sentiment Analysis, with a focus on the transition from traditional fusion methods to advanced Transformer-based models. We then survey the landscape of Parameter-Efficient Fine-Tuning techniques, which are central to addressing the computational challenges of large models. Finally, we examine the principles of Mixture-of-Experts models, the architectural paradigm that serves as a key point of comparison for our work.

### 2.1 Multimodal Sentiment Analysis

Multimodal Sentiment Analysis (MSA) aims to infer sentiment by interpreting and integrating information from multiple communication channels, typically language, acoustics, and vision (Baltrušaitis et al., 2018). The primary challenge in MSA lies in effectively modeling both the intra-modal dynamics within each modality and the complex inter-modal interactions between them. Early research in this field primarily focused on designing effective feature fusion strategies. These strategies are often categorized into early (feature-level) fusion, where multimodal features are concatenated at the input stage before being fed into a classifier, and late (decision-level) fusion, where predictions from unimodal classifiers are combined, for instance, through a voting scheme (Poria et al., 2017). While straightforward, these methods often struggle to capture the nuanced, non-linear dependencies between modalities. To address this, more sophisticated fusion techniques like tensor-based fusion were proposed to model higher-order correlations (Zadeh et al., 2017).

The advent of deep learning, particularly attention mechanisms, has significantly advanced the field by enabling models to learn the intricate interplay between modalities dynamically. The introduction of the Transformer architecture (Devlin et al., 2019), with its powerful sequence modeling and cross-representational learning capabilities, marked a revolutionary breakthrough for MSA. A seminal work in this direction, the Multimodal Transformer (MulT) (Tsai et al., 2019), introduced cross-modal attention mechanisms that allow representation sequences from different modalities to attend to each other, effectively addressing the challenge of temporal asynchronicity in unaligned multimodal data. Building on this foundation, subsequent research has focused on adapting powerful Pre-trained Language Models (PLMs) for multimodal tasks. For example, MAG-BERT (Multimodal Adaptation Gate for BERT) (Sun et al., 2020) designed a sophisticated gating unit to adaptively modulate the influence of visual and acoustic features on the internal representations of BERT. Other works, such as MISA (Hazarika et al., 2020), have explored disentangling multimodal representations into modality-invariant and modality-specific subspaces to achieve more robust fusion. These works demonstrated the efficacy of fine-grained multimodal fusion in the PLM era but typically rely on full fine-tuning, which incurs substantial computational costs and motivates the need for more efficient adaptation methods.

## 2.2 Parameter-Efficient Fine-Tuning

Parameter-Efficient Fine-Tuning (PEFT) has emerged as a dominant paradigm to adapt large pre-trained models to downstream tasks without incurring the prohibitive costs of full fine-tuning. The core principle of PEFT is to freeze the vast majority of the pre-trained parameters and optimize only a small number of newly added or selected parameters, thereby drastically reducing the memory and storage footprint.

Adapter Tuning (Houlsby et al., 2019) is a foundational PEFT technique that has shown remarkable success. It involves inserting small, bottleneck-structured neural networks, termed "adapters," within each layer of a pre-trained Transformer. During fine-tuning, only the parameters of these adapter modules are updated. This approach isolates task-specific knowledge within the adapters, preserving the general knowledge of the pre-trained model and enabling efficient multi-task learning.

Another highly influential PEFT method is LoRA (Low-Rank Adaptation) (Hu et al., 2022). LoRA is based on the hypothesis that the change in model weights during task adaptation has a low intrinsic rank. Consequently, it approximates this weight update by training two small, low-rank matrices and injects their product into the original weight matrices of the pre-trained model. In addition to these module-based approaches, other PEFT families have emerged, such as prompt-based methods. Prefix-Tuning (Li & Liang, 2021) and Prompt-Tuning (Lester et al., 2021) prepend a small set of trainable continuous vectors (or "prompts") to the input sequence, steering the behavior of the frozen PLM without modifying its internal weights. Our work leverages the modularity of both LoRA and Adapters, using the former to adapt the core BERT backbone and the latter as the fundamental building blocks for our expert networks.

## 2.3 Mixture-of-Experts Models

The Mixture-of-Experts (MoE) model is a powerful and scalable architecture for conditional computation. The central idea is to decompose a large network into multiple smaller "expert" subnetworks and use a "gating network" or "router" to dynamically and sparsely select a subset of these experts for each input sample. This allows for a massive increase in the model's total parameter count (and thus its capacity) without a proportional increase in computational cost (FLOPs).

The Sparsely-Gated Mixture-of-Experts layer (Shazeer et al., 2017) laid the theoretical groundwork for modern MoE models. This work introduced a differentiable gating network that produces a probability distribution over all experts for each input and routes the information only to the top-K highest-scoring experts. To prevent the routing strategy from collapsing to using only a few popular experts, this work also introduced an auxiliary Load Balancing Loss, which encourages the gating network to distribute inputs roughly uniformly across all experts. Subsequent works, such as the Switch Transformer (Fedus et al., 2022), have further simplified and scaled the MoE design to models with over a trillion parameters, demonstrating its effectiveness in large-scale language modeling.

In the context of MSA, the Mixture of Multimodal Adapters (MMA) framework (Chen et al., 2025), which serves as the baseline for our study, was the first to combine the MoE concept with Adapters for this specific task. It validated the feasibility of using lightweight Adapters as experts in a multimodal PEFT setting. However, its use of a "flat" routing mechanism and homogeneous experts leaves room for improvement in terms of interpretability and specialized expert capacity, challenges that our present work aims to address by replacing this paradigm with a novel serial, intra-layer fusion approach.

### III. METHODOLOGY

Building upon the foundations of parameter-efficient fine-tuning and multimodal fusion, this chapter details the architecture and mechanisms of our proposed model. As established in the introduction, while frameworks like the Mixture of Multimodal Adapters (MMA) (Chen et al., 2025) have pioneered parameter-efficient approaches, they exhibit limitations in their static fusion, flat routing, and homogeneous expert design. To overcome these challenges, we introduce the Progressive Gated-Fusion Network (PGF-Net) a novel architecture designed for deep, dynamic, and interpretable multimodal sentiment analysis.

The cornerstone of PGF-Net is the principle of Progressive Intra-Layer Fusion. This paradigm shifts away from a single, discrete fusion event towards a continuous, hierarchical process. Non-linguistic information from audio and visual streams is progressively injected into the textual representation as it flows through the deep layers of a Transformer-based encoder. This ensures that fusion is not an afterthought but an integral part of the representation learning process, enabling the model to capture more complex, context-dependent inter-modal dynamics.

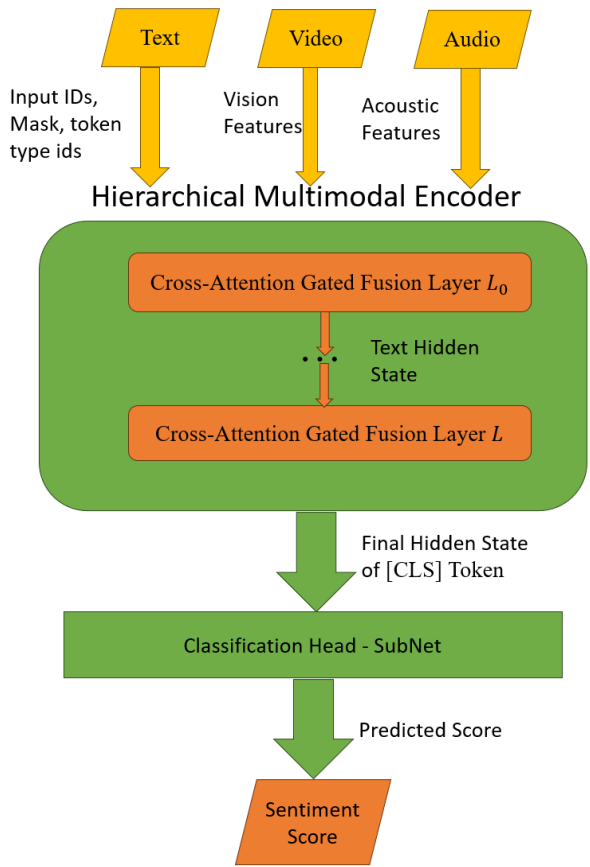

Fig. 1.    The details of the PGF-Net

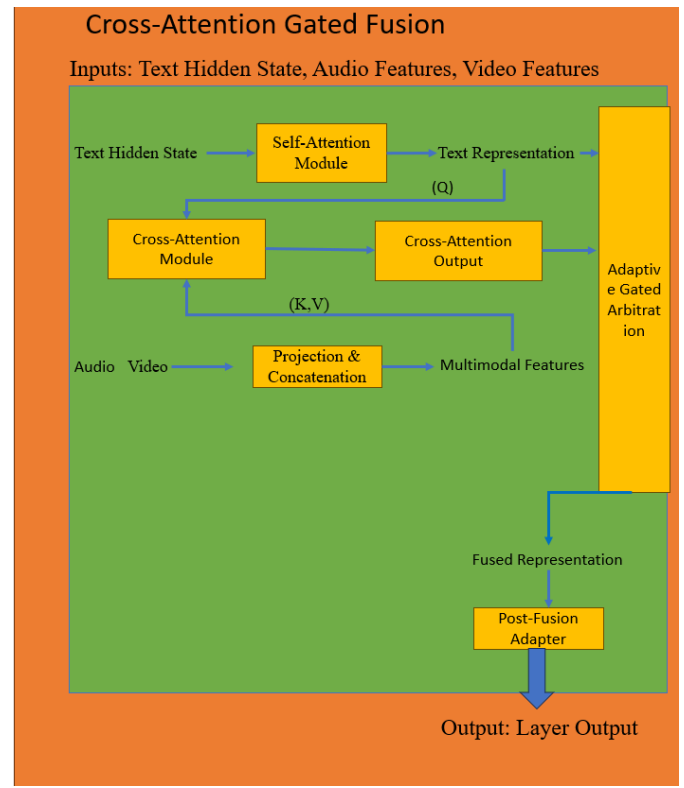

Fig. 2.    The details of the Cross-Attention Gated Fusion Layer

### 3.1  Overall Architecture of PGF-Net

The overview of our proposed PGF-Net is illustrated in Figure 1. It comprises three components: a Multimodal Input Representation, a Hierarchical Multimodal Encoder, and a lightweight Regression Head.

3.1.1 Input Representation

The model consumes three time-aligned sequences provided by the dataset pipeline: a tokenized text sequence T, an acoustic sequence A, and a visual sequence V:

1. Textual Representation: The raw textual sequence (T) is processed using the tokenizer. Each sentence is converted into a sequence of tokens, which are then mapped to numerical IDs. The final textual input consists of input_ids, attention_mask, and token_type_ids, which are fed into the embedding layer of the hierarchical encoder.

2. The acoustic modality (A) is provided as a time-aligned sequence of preprocessed acoustic descriptors from the dataset pipeline, capturing prosodic and spectral cues (e.g., pitch-, energy-, and spectrum-related information). We use these descriptors directly as inputs to the encoder.

3. Visual Representation: The visual modality (V) is provided as a time-aligned sequence of preprocessed visual descriptors from the dataset pipeline—typically reflecting facial actions, head/eye pose, and related cues. We use these descriptors directly as inputs to the encoder; the exact extraction settings and dimensionality are reported in chapter 4.

### 3.1.2 Hierarchical Multimodal Encoder

This is the heart of our model. The encoder is a Transformer-style stack whose standard self-attention and feed-forward sublayers are preserved. Starting from a configurable depth $L_0$, each layer is augmented with our Cross-Attention–Gated Fusion block, which integrates three components: (i) a cross-attention module that retrieves non-linguistic cues from an audio–visual feature bank, (ii) an adaptive gating mechanism that balances linguistic and non-linguistic information, and (iii) a post-fusion adapter for task-specific refinement. Layers with $l < L_0$ operate as vanilla text-only Transformer layers (no fusion); for all $l \geq L_0$ the same in-layer fusion sequence is executed. We refer to the encoder as *hierarchical* because this intra-layer fusion is repeated across multiple depths, progressively injecting audio-visual context throughout the stack rather than performing a single late-fusion step. Figure 2 illustrates the details of the Cross-Attention Gated Fusion Layer

### 3.1.3 Classification Head

The final hidden state vector corresponding to the [CLS] token from the last encoder layer (Layer L) is utilized as the comprehensive multimodal representation for the entire input. This feature vector is then passed to a lightweight feed-forward network, implemented as the SubNet class. The SubNet consists of a stack of linear layers with ReLU activation and dropout, which regresses the 768-dimensional feature vector into a final, one-dimensional sentiment score.

### 3.2 The Core Module: Cross-Attention Gated Fusion Layer

The primary innovation of our work is encapsulated within the design of the fusion-stage Cross-Attention Gated Fusion Layers. This module seamlessly integrates unimodal contextualization and cross-modal integration within a single layer's computation. The detailed data flow is depicted in Figure 2.

For any given fusion layer $l$ (where $l \geq N$), the computation proceeds through four synergistic steps, taking the textual hidden state from the previous layer, $H_{l-1}$, as its primary input.

### 3.2.1 Unimodal Contextualization via Self-Attention

First, the incoming text representation $H_{l-1}$ is processed by a standard multi-head self-attention mechanism, identical to that in the original Transformer architecture (Devlin et al., 2019). This step captures the intricate contextual dependencies between tokens within the text stream, producing a contextually-aware text representation, $H_{text}$.

$$H_{text} = \text{Self-Attention}(H_{l-1}) \quad (1)$$

This ensures that a deep understanding of the linguistic context is established before any cross-modal information is introduced.

### 3.2.2 Cross-Modal Information Extraction via Cross-Attention

In parallel, the non-linguistic modalities are prepared for fusion. The raw acoustic ($X_a$) and visual ($X_v$) feature sequences are first projected into the same hidden dimension as the text representation ($D_{bert}$) via separate linear layers. These projected features are then concatenated along the sequence dimension to form a unified multimodal feature bank, $X_{av}$.

$$X'_a = X_a W_a + b_a \; ; \; X'_v = X_v W_v + b_v \quad (2)$$

$$H_{av} = \text{Concat}(X'a, X'v) \quad (3)$$

The key step of inter-modal interaction is performed by a cross-attention module, a mechanism central to successful multimodal Transformers like MulT (Tsai et al., 2019). Here, the text representation $H_{text}$ acts as the Query, while the multimodal feature bank $H_{av}$ serves as both the Key and Value.

$$H_{cross} = \text{Cross-Attention}(Q=H_{text}, K=H_{av}, V = H_{av}) \quad (4)$$

This process can be interpreted as an information extraction routine: each token in the text representation is empowered to actively "query" the entire repository of audio-visual information and extract the most relevant features. The result, $H_{cross}$, is a new representation of the text sequence, where each token is now enriched with pertinent non-linguistic context.

### 3.2.3 Adaptive Gated Arbitration

A critical challenge in multimodal fusion is preventing informative linguistic features from being overwhelmed by less relevant or noisy non-linguistic signals. To address this, we introduce an Adaptive Gated Arbitration mechanism, inspired by the gating principles in units like GRU (Cho et al., 2014). This gate acts as a dynamic arbitrator, learning to control the flow of information. It computes a fusion coefficient, g, based on both the original text representation and the cross-modally enriched representation.

$$g = \sigma(W_g[H_{text}; H_{cross}] + b_g) \quad (5)$$

$$H_{fused} = g \odot H_{text} + (1 - g) \odot H_{cross} \quad (6)$$

Here, [;] denotes concatenation, σ is the sigmoid function, and ⊙ represents element-wise multiplication. This mechanism allows the model to learn a fine-grained, context-dependent balance, deciding for each feature dimension whether to preserve the original linguistic information ($g \to 1$) or to emphasize the newly incorporated multimodal context ($g \to 0$).

### 3.2.4 Post-Fusion Refinement with Adapters

Finally, the arbitrated fused representation, $H_{fused}$, is passed through a Post-Fusion Adapter. This module, consistent with the principles of Adapter-Tuning (Houlsby et al., 2019), is a small, bottleneck-structured network. Its function is to perform a final, non-linear transformation on the already-fused representation. This step acts as a refinement stage, allowing the model to learn specialized transformations for the complex, fused feature space, thereby enhancing its expressive power without significantly increasing the parameter count. The output of the adapter, $H_l$, becomes the final output of the layer.

### 3.3 Parameter-Efficient Fine-Tuning Strategy

Our framework is designed from the ground up for parameter efficiency. This is achieved by synergistically combining two leading PEFT techniques:

1. LoRA (Low-Rank Adaptation) (Hu et al., 2022): The entire BERT backbone is adapted using LoRA. By freezing the original weights and only training low-rank decomposition matrices for the query and value projections in the attention mechanisms, we can adapt the large model to the specifics of the sentiment analysis task with minimal trainable parameters.
2. Adapter Tuning (Houlsby et al., 2019): The Post-Fusion Adapter modules introduced in each fusion layer are themselves a form of PEFT. These lightweight modules contain a very small number of parameters compared to the main model but provide crucial task-specific modeling capacity at the point of fusion.

This dual strategy of "global adaptation via LoRA and local refinement via Adapters" ensures that our PGF-Net is both powerful and computationally frugal, making it suitable for deployment in environments with limited resources.

### 3.4 Prediction and Optimization Objective

After the input propagates through all L layers of the encoder, the final hidden state of the [CLS] token, $H_L^{[CLS]}$, is extracted as the ultimate sentence-level multimodal representation. This vector is then fed into a classification head (SubNet), which consists of a simple multi-layer perceptron, to produce the final scalar sentiment score.

The model is trained end-to-end by minimizing the Mean Absolute Error (L1 Loss) between the predicted scores and the ground-truth labels. This loss function is chosen for its robustness to outliers, which are common in subjective sentiment annotations.

## IV. EXPERIMENTS

### 4.1 Datasets

Our experiments were conducted on two widely-used benchmark datasets in multimodal sentiment analysis:

- CMU-MOSI (Zadeh et al., 2016): As one of the most classic benchmarks in the field, CMU-MOSI comprises 2,199 short video clips expressing singular sentiment, extracted from YouTube. Each clip is annotated with data from three modalities text, vision, and audio and includes a single human sentiment annotation on a scale from [-3, +3], representing sentiment intensity from highly negative to highly positive.
- CMU-MOSEI (Zadeh et al., 2018): As a large-scale extension of MOSI, CMU-MOSEI is one of the largest sentiment analysis datasets available today. It contains 23,453 video sentence segments from over 1,000 unique speakers, covering a broader range of topics and more natural conversational settings. Its annotation scheme is identical to that of MOSI. Due to its scale and complexity, MOSEI presents a greater challenge to the robustness and generalization capabilities of models.

We adhere to the standard training, validation, and test set splits to ensure a fair comparison with previous works.

### 4.2 Evaluation Metrics and Implementation Details

#### 4.2.1 Evaluation Metrics

To provide a comprehensive assessment of model performance, we adopted a suite of metrics consistent with prior research (Chen et al., 2025) (Hazarika et al., 2020):

- Mean Absolute Error (MAE): Measures the average absolute difference between predicted and true sentiment scores. Lower is better.
- Pearson Correlation Coefficient (Corr): Measures the linear correlation between predicted and true values. Higher is better.
- Binary Accuracy (Acc-2): Accuracy calculated after classifying sentiment scores as positive (non-negative) versus negative.
- 7-Class Accuracy (Acc-7): Accuracy calculated after rounding sentiment scores to the nearest integer (in the range [-3, 3]).
- F1-Score: The harmonic mean of precision and recall for the binary classification task.
- Trainable Parameters (TP): Measures the parameter efficiency of the model, reported in millions (M). Lower is better.

TABLE I. EXPERIMENTAL RESULTS ON MOSI DATASET

| Methods | Param | MAE | Corr | ACC-7 | ACC-2 | F1 |
|---|---|---|---|---|---|---|
| TFN | - | 0.901 | 0.698 | 34.9 | 80.8 | 80.7 |
| LMF | - | 0.917 | 0.695 | 33.2 | 82.5 | 82.4 |
| MulT | - | 0.861 | 0.711 | - | 84.1 | 83.9 |
| MISA | 110.6M | 0.783 | 0.761 | 42.3 | 83.4 | 83.6 |
| MAG | 110.9M | 0.712 | 0.796 | - | 86.1 | 86.0 |
| Self-MM | 109.7M | 0.713 | 0.798 | - | 86.0 | 86.0 |
| CubeMLP | 110.6M | 0.770 | 0.767 | 45.5 | 85.6 | 85.5 |
| ConFEDE | 129.7M | 0.742 | 0.784 | 42.3 | 85.5 | 85.5 |
| AcFormer | 130.2M | 0.715 | 0.794 | 44.2 | 85.4 | 85.7 |
| MMA | 5.7M | 0.693 | 0.803 | 46.9 | 86.4 | 86.4 |
| PGF-Net | **3.09M** | **0.691** | **0.809** | **49.4** | **86.8** | **86.9** |

### 4.2.2 Implementation Details

Our PGF-Net model was implemented using the PyTorch framework, with the pre-trained bert-base-uncased model (Devlin et al., 2019) loaded from the HuggingFace Transformers library. All experiments were conducted on a single NVIDIA 4090 GPU. Key hyperparameters were set as follows: the AdamW optimizer was used with an initial learning rate of 1e-3 and a weight decay of 1e-4. The LR is decayed by StepLR with step_size=9 and γ=0.1. The batch size was set to 128, and the model was trained for a maximum of 20 epochs with an early stopping strategy (patience=10). For our PEFT strategy, the rank for LoRA (Hu et al., 2022) was set to 32, and the bottleneck dimension for the Post-Fusion Adapter (Houlsby et al., 2019) was 64. Within the PGF-Net architecture, fusion was configured to begin at the very first layer of the BERT encoder (index 0) to maximize the depth of inter-modal interaction.

### 4.3 Comparisons with Other Methods

To validate the effectiveness of our proposed PGF-Net, we conducted comprehensive experiments on the MOSI and MOSEI datasets, comparing its performance against various mainstream methods, including TFN (Zadeh et al., 2017), LMF (Liu et al., 2018), MuLT (Tsai et al., 2019), MISA (Hazarika et al., 2020), MAG (Sun et al., 2020), Self-MM (Yuan et al., 2021), CubeMLP (Mao et al., 2022), ConFEDE (Liu et al., 2023), Acformer (Yang et al., 2024), and MMA (Chen et al., 2025) (as detailed in Table I and II). The analysis reveals that on the MOSI dataset, PGF-Net delivers the best performance across all evaluation metrics. On the larger and more complex MOSEI dataset, PGF-Net demonstrates a highly competitive and nuanced performance profile, underscoring the specific strengths of our architecture. Notably, our model achieves a state-of-the-art

F1-Score of 86.1 and the highest 7-class accuracy of 58.5. This superiority in classification-based metrics strongly suggests that our progressive, layer-by-layer fusion mechanism is particularly effective at capturing the subtle, complex emotional cues required for fine-grained sentiment categorization.

While its MAE (0.525) and Pearson Correlation (0.785) do not set new records, they remain on par with top-performing models like ConFEDE and AcFormer. It is crucial, however, to contextualize this performance with respect to parameter efficiency. PGF-Net achieves these results with only 4.72M trainable parameters, approximately half that of MMA and a small fraction of the other leading methods. This trade-off suggests that while our model might not perfectly align with the linear sentiment scale as measured by Corr on this highly complex dataset, it provides a state-of-the-art balance between nuanced classification accuracy and exceptional computational efficiency. This makes PGF-Net a particularly practical and robust framework, especially for applications where both high performance and resource economy are critical.

TABLE II. EXPERIMENTAL RESULTS ON MOSEI DATASET

| Methods | Param | MAE | Corr | ACC-7 | ACC-2 | F1 |
|---|---|---|---|---|---|---|
| TFN | - | 0.593 | 0.700 | 50.2 | 82.5 | 82.1 |
| LMF | - | 0.623 | 0.677 | 48.0 | 82.0 | 82.1 |
| MulT | - | - | - | - | 83.5 | 82.9 |
| MISA | 47.1M | 0555 | 0.756 | 52.2 | 85.5 | 85.3 |
| MAG | 111.8M | - | - | - | 84.7 | 84.5 |
| Self-MM | 109.7M | 0.530 | 0.765 | - | 85.2 | 85.3 |
| CubeMLP | 110.6M | 0.529 | 0.760 | 54.9 | 85.1 | 84.5 |
| ConFEDE | 137.0M | **0.522** | 0.780 | 54.9 | 85.8 | 85.8 |
| AcFormer | 130.1M | 0.531 | **0.786** | 54.7 | 86.5 | 85.8 |
| MMA | 8.1M | 0.529 | 0.766 | 55.2 | 85.7 | 85.7 |
| PGF-Net | **4.72M** | 0.525 | 0.785 | **58.5** | **86.1** | **86.1** |

### 4.4 Ablation Study

To rigorously dissect the architecture of our proposed PGF-Net and quantify the individual contributions of its core components, we conducted a comprehensive ablation study on the MOSI dataset. By systematically removing or deactivating key modules, we can analyze their impact on the model's overall performance and efficiency. The results of this study are presented in Table III.

An analysis of the results in Table III yields several critical insights into our model's design.

TABLE III. ABLATION STUDY ON MOSI DATASET

| Methods | Param | MAE | Corr | ACC-7 | ACC-2 | F1 |
|---|---|---|---|---|---|---|
| PGF-Net | 3.09M | 0.691 | 0.809 | 49.4 | 86.8 | 86.9 |
| w/o CA (Cross-Attention) | 3.09M | 0.725 | 0.783 | 45.8 | 84.5 | 84.6 |
| w/o Gate (Gating Mechanism) | 3.09M | 0.710 | 0.796 | 47.5 | 85.7 | 85.8 |
| w/o Refiner (Post-Fusion Adapter) | 5.12M | 0.699 | 0.802 | 48.7 | 86.5 | 86.6 |
| w/o CA & w/o Gate | 3.09M | 0.731 | 0.779 | 45.2 | 84.0 | 84.0 |
| w/o CA & w/o Refiner | 5.7M | 0.729 | 0.782 | 45.4 | 84.1 | 84.1 |
| w/o Gate & w/o Refiner | 5.12M | 0.721 | 0.708 | 47.6 | 86.0 | 86.1 |

#### 4.4.1 Evaluation Metrics

The most pronounced performance degradation across all metrics occurred upon the removal of the Cross-Attention module (w/o CA). This variant exhibited a substantial increase in MAE by 0.034 and a decrease in Corr by 0.026. Most notably, the fine-grained 7-class accuracy (Acc-7) experienced a significant decline of 3.6 percentage points. This performance drop was further exacerbated when CA was removed in conjunction with other modules (w/o CA & Gate, w/o CA & Refiner), confirming its foundational role.

This observation strongly validates that the Cross-Attention mechanism is the cornerstone of effective modality fusion in our architecture. Its removal forces the model to resort to rudimentary fusion strategies (e.g., simple addition or concatenation), which are incapable of dynamically extracting contextually relevant information. Without the ability for the text representation to selectively "query" and integrate salient cues from the audio-visual stream, the model's capacity for deep inter-modal reasoning is severely compromised. This deficiency is most evident in the sharp decline of the Acc-7 score, which is highly sensitive to the model's ability to interpret nuanced emotional expressions.

### 4.4.2 Efficacy of the Gated Arbitration Mechanism (Gate)

The ablation of the Fusion Gate (w/o Gate) also resulted in a consistent, albeit less severe, decline in performance. MAE increased by 0.019, while Acc-7 dropped by 1.9 points. This demonstrates that the adaptive gating mechanism serves as a crucial information flow regulator.

Without the gate, the model is forced into a naive summation of the textual representation and the cross-attention output. Such an approach is suboptimal, as it may either allow noisy or irrelevant cross-modal signals to corrupt the original linguistic semanticsor, conversely, dilute potent non-verbal cues with equally weighted textual information. The gating mechanism, by learning a dynamic arbitration coefficient, intelligently balances the preservation of unimodal context with the integration of cross-modal enrichment. This leads to a more robust and refined fusion process, the efficacy of which is clearly substantiated by the experimental results.

### 4.4.3 The Dual Role of the Post-Fusion Refiner (Adapter)

The removal of the Post-Fusion Adapter (w/o Refiner) reveals two significant insights. First, it leads to a slight but consistent drop in performance metrics (e.g., a 0.7-point decrease in Acc-7), confirming its utility in refining the fused representation. Second, and more critically, its removal causes the number of trainable parameters to surge from 3.09M to 5.12M. This is because, in the w/o Refiner variant, additional backbone layers are unfrozen to ensure convergence, whereas in other settings only the adapters and gating modules remain trainable.

This underscores the dual role of the adapter:

1. Performance Refiner: The adapter applies a final, task-oriented non-linear transformation to the already-fused multimodal features. This allows the model to capture higher-order feature interactions, providing a "last-mile" enhancement to predictive accuracy.

2. Parameter Efficiency Enabler: This experiment provides compelling inverse evidence for our PEFT strategy. The adapter is a cornerstone of the model's efficiency. To achieve convergence without this lightweight module, a significantly larger portion of the BERT backbone's parameters must be unfrozen and fine-tuned, leading to a substantial increase in the trainable parameter count. Therefore, the Post-Fusion Adapter is not merely a performance booster but a fundamental pillar of our parameter-efficient design.

### 4.4.4 Analysis of Synergistic Effects

When multiple components were removed simultaneously (e.g., w/o CA & Gate), the resulting performance degradation was consistently greater than the sum of the individual ablations. For instance, the Acc-7 score for w/o CA & Gate (45.2%) is lower than that for both w/o CA (45.8%) and w/o Gate (47.5%).

This clearly indicates a strong synergistic effect among the proposed modules. They do not function as simple, additive enhancements but as integral parts of a cohesive, integrated system. The Cross-Attention module "extracts" information, the Fusion Gate "arbitrates" it, and the Post-Fusion Adapter "refines" it. It is this logically sound and deeply integrated design that collectively contributes to the superior performance and remarkable parameter

## V. CONCLUSIONS

In this paper, we addressed the dual challenges of achieving effective fusion and high parameter efficiency in Multimodal Sentiment Analysis. We identified critical limitations in existing parameter-efficient frameworks that utilize parallel expert modules, namely the potential for insufficient inter-modal interaction and the adoption of rigid fusion strategies. To overcome these issues, we proposed and developed a novel architecture, the Progressive Gated-Fusion Network (PGF-Net).Our work's core contribution is the introduction of a progressive intra-layer fusion paradigm, which fundamentally shifts from a parallel, "divide-and-conquer" strategy to a serial, "progressive enhancement" model. We demonstrated how the synergistic combination of cross-attention for information extraction, an adaptive gate for information arbitration, and a post-fusion adapter for feature refinement within each encoder layer facilitates a deeper and more context-aware integration of multimodal signals. The empirical results presented in this study robustly validate the efficacy of our approach. On the widely-used CMU-MOSI benchmark, PGF-Net not only achieved performance on par with or exceeding state-of-the-art methods but did so with exceptional parameter efficiency, requiring only 3.1 million trainable parameters. The significant improvement in 7-class accuracy particularly underscores our model's enhanced capability to discern nuanced emotional states, a direct benefit of its deep fusion mechanism.

While PGF-Net has demonstrated considerable promise, our work also opens up several avenues for future investigation. Our analysis on the more complex CMU-MOSEI dataset indicates that future work should focus on enhancing the model's robustness for "in-the-wild" scenarios. This could involve integrating more advanced noise-suppression techniques or exploring more sophisticated gating mechanisms adept at handling longer sequences and dynamic emotional shifts. Another promising direction is the exploration of heterogeneous and dynamic adapters, where the structure or function of the adapter could vary with network depth or adapt based on the input, potentially yielding further performance gains. Finally, the progressive fusion paradigm proposed in PGF-Net is generalizable. We plan to extend our framework to other challenging multimodal tasks, such as Emotion Recognition in Conversation (ERC) and Visual Question

Answering (VQA), to further assess its versatility and effectiveness across different domains.

In summary, this research validates that a serial, progressive fusion architecture offers an effective and highly efficient path toward advanced multimodal understanding, presenting a compelling alternative to traditional parallel expert models.

## VI. ACKNOWLEDGEMENT

This research was supported by Horizon 2020 grant: Exchanges for SPEech ReseArch aNd TechnOlogies (ESPERANTO), Grant agreement ID: 101007666.

## REFERENCES

Baltrušaitis, T., Ahuja, C., & Morency, L. P. (2018). Multimodal machine learning: A survey and taxonomy. IEEE Transactions on Pattern Analysis and Machine Intelligence, 41(2), 423–443.

Liu, B. (2012). Sentiment analysis and opinion mining. Morgan & Claypool Publishers.

Devlin, J., Chang, M.-W., Lee, K., & Toutanova, K. (2019). BERT: Pre-training of deep bidirectional transformers for language understanding. In Proceedings of NAACL-HLT 2019.

Tsai, Y.-H. H., Bai, S., Liang, P. P., Kolter, J. Z., Morency, L.-P., & Salakhutdinov, R. (2019). Multimodal transformer for unaligned multimodal language sequences. In Proceedings of ACL 2019.

Sun, Z., Sar-Goldman, Y., & Radke, R. J. (2020). Improving multimodal sentiment analysis via multimodal adaptation gate. In Findings of EMNLP 2020.

Houlsby, N., Giurgiu, A., Jastrzebski, S., Morrone, B., De Laroussilhe, Q., Gesmundo, A., Attariyan, M., & Gelly, S. (2019). Parameter-efficient transfer learning for NLP. In Proceedings of ICML 2019.

Hu, E. J., Shen, Y., Wallis, P., Allen-Zhu, Z., Li, Y., Wang, S., Wang, L., & Chen, W. (2022). LoRA: Low-rank adaptation of large language models. In Proceedings of ICLR 2022.

Shazeer, N., Mirhoseini, A., Maziarz, K., Davis, A., Le, Q., Hinton, G., & Dean, J. (2017). Outrageously large neural networks: The sparsely-gated mixture-of-experts layer. In Proceedings of ICLR 2017.

Chen, K., Ben, H., Wang, S., Tang, S., & Hao, Y. (2025). Mixture of multimodal adapters for parameter-efficient multimodal sentiment analysis. In Proceedings of NAACL-HLT 2025.

Cho, K., Van Merriënboer, B., Gulcehre, C., Bahdanau, D., Bougares, F., Schwenk, H., & Bengio, Y. (2014). Learning phrase representations using RNN encoder–decoder for statistical machine translation. In Proceedings of EMNLP 2014.

Fedus, W., Zoph, B., & Shazeer, N. (2022). Switch transformers: Scaling to trillion parameter models with simple and efficient sparsity. Journal of Machine Learning Research, 23(1), 1–39.

Poria, S., Cambria, E., Bajpai, R., & Hussain, A. (2017). A review of affective computing: From unimodal analysis to multimodal fusion. Information Fusion, 37, 98–125.

Zadeh, A., Chen, M., Poria, S., Cambria, E., & Morency, L.-P. (2017). Tensor fusion network for multimodal sentiment analysis. In Proceedings of EMNLP 2017.

Hazarika, D., Zimmermann, R., & Poria, S. (2020). MISA: Modality-invariant and -specific representations for multimodal sentiment analysis. In Proceedings of ACM MM 2020.

Li, X. L., & Liang, P. (2021). Prefix-tuning: Optimizing continuous prompts for generation. In Proceedings of ACL 2021.

Lester, B., Al-Rfou, R., & Constant, N. (2021). The power of scale for parameter-efficient prompt tuning. In Proceedings of EMNLP 2021.

Zadeh, A. B., Zellers, R. G., Pincus, E., & Morency, L.-P. (2016). MOSI: Multimodal corpus of sentiment intensity and subjectivity analysis in online opinion videos. In Proceedings of ACL 2016 (Short Papers).

Zadeh, A., Liang, P. P., Poria, S., Cambria, E., & Morency, L.-P. (2018). CMU-MOSEI: A multimodal language dataset for sentiment and emotion analysis in the wild. In Proceedings of ACL 2018.)

Liu, Z., Shen, Y., Lakshminarasimhan, V. B., Liang, P. P., Zadeh, A., & Morency, L.-P. (2018). Efficient low-rank multimodal fusion with modality-specific factors. In Proceedings of ACL 2018.

Yuan, J., Kui, H., Yang, J., Xin, M., Liu, Y., & Wang, L. (2021). Self-MM: A self-supervised learning method for multimodal sentiment analysis with incomplete modalities. In Proceedings of AAAI 2021.

Mao, L., Mai, Z., Yang, Z., Zhang, Y., Kui, H., Zhao, J., & Wang, L. (2022). Are we all good at playing Tetris? A study on the biases of vision–language models in multimodal sentiment analysis. In Findings of ACL 2022.

Liu, Y., Zhang, L., Li, G., Yong, T., Liu, Y., Yao, B., & Li, D. (2023). ConFEDE: Contrastive fusion encoder with dynamic feature-enhanced embeddings for multimodal sentiment analysis. IEEE Transactions on Multimedia, 25, 6967–6978.

Yang, Y., Li, C., Chen, Z., Lin, L., & Luo, P. (2024). AcFormer: Adapter-centric transformer for parameter-efficient transfer tuning in multimodal sentiment analysis. In Proceedings of The Web Conference 2024.